\documentclass[11pt,a4paper]{article}
\usepackage[hyperref]{AACL-IJCNLP2020}
\usepackage{times}
\usepackage{latexsym}

\usepackage{microtype}

\usepackage{helvet}
\usepackage{courier}
\usepackage{url}
\usepackage{graphicx}
\usepackage{mathptmx}
\usepackage[hyphenbreaks]{breakurl}
\usepackage{xcolor}
\usepackage{bibentry}
\usepackage{stfloats}
\usepackage{subfig}
\usepackage{multirow}
\usepackage{amssymb}
\usepackage{pifont}
\usepackage{amsmath}
\allowdisplaybreaks
\usepackage{array}

\DeclareMathAlphabet{\mathcal}{OMS}{cmsy}{m}{n}
\definecolor{mypink}{rgb}{1., 0, 0}
\usepackage{enumitem}
\usepackage{stmaryrd}

\usepackage{algorithm}
\usepackage{algorithmic}
\renewcommand{\algorithmiccomment}[1]{\bgroup\hfill~#1\egroup}
\usepackage{footnote}

\aclfinalcopy 

\title{MaP: A Matrix-based Prediction Approach to Improve Span Extraction in Machine Reading Comprehension}

\author{Huaishao Luo$^{1}$\thanks{~~This work was done during the first author's internship at Microsoft}~, Yu Shi$^2$, Ming Gong$^3$, Linjun Shou$^3$, Tianrui Li$^{1}$ \\
  $^1$School of Information Science and Technology, Southwest Jiaotong University \\
  $^2$Microsoft Cognitive Services Research Group \\
  $^3$Microsoft STCA NLP Group \\
  {\tt huaishaoluo@gmail.com, trli@swjtu.edu.cn} \\
  {\tt \{yushi,migon,lisho\}@microsoft.com}
  } 

\date{}

\begin{document}
	\maketitle
	\begin{abstract}
		Span extraction is an essential problem in machine reading comprehension. Most of the existing algorithms predict the start and end positions of an answer span in the given corresponding context by generating two probability vectors. In this paper, we propose a novel approach that extends the probability vector to a probability matrix. Such a matrix can cover more start-end position pairs. Precisely, to each possible start index, the method always generates an end probability vector. Besides, we propose a sampling-based training strategy to address the computational cost and memory issue in the matrix training phase. We evaluate our method on SQuAD 1.1 and three other question answering benchmarks. Leveraging the most competitive models BERT and BiDAF as the backbone, our proposed approach can get consistent improvements in all datasets, demonstrating the effectiveness of the proposed method.
	\end{abstract}

	\section{Introduction}
	\label{sec_introduction}
	Machine reading comprehension (MRC), which requires the machine to answer comprehension questions based on the given passage of text, has been studied extensively in the past decades \cite{Liu2019}. Due to the increase of various large-scale datasets (e.g., SQuAD \cite{Rajpurkar2016} and MS MARCO \cite{Nguyen2016}), and the enhancement of pre-trained models (e.g., ELMo \cite{Peters2018}, BERT \cite{Devlin2019}, and XLNet \cite{Yang2019a}), remarkable advancements have been made recently in this area. Among various MRC tasks, span extraction is one of the essential tasks. Given the context and question, the span extraction task is to extract a span of the most plausible text from the corresponding context as a candidate answer. Although there exist unanswerable cases beyond the span extraction, the span-based task is still fundamental and significant in the MRC field.
	\begin{figure}[tbp]
		\centering
		\includegraphics[width=3.2in]{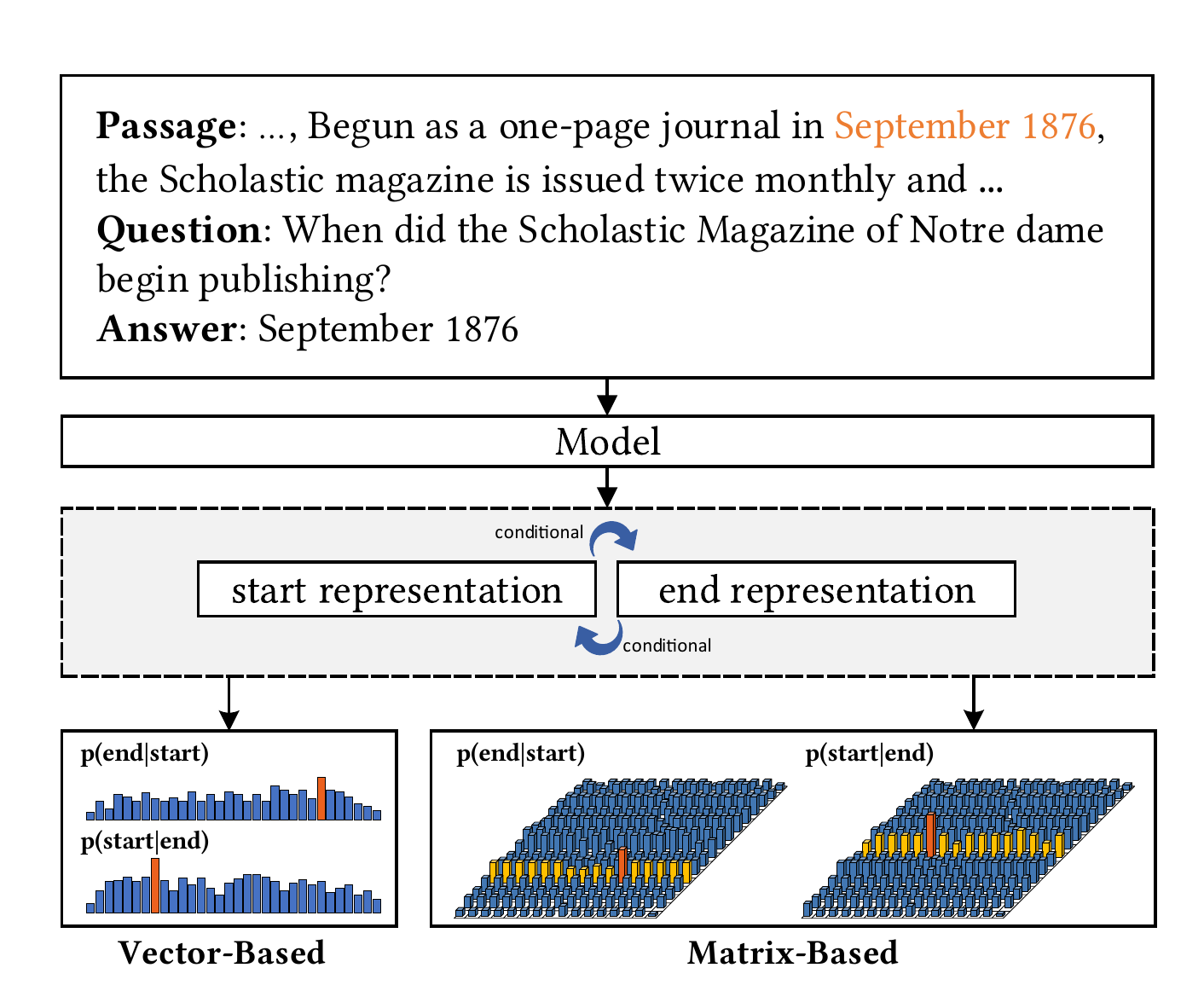}
		\caption{An illustration of a machine reading comprehension framework. Most of previous works are vector-based approaches shown as the left part. Our matrix-based conditional approach is shown in the right part. In our setting, every start (or end) position has an end (or start) probability vector, which leads that the output probabilities is a matrix (best seen in color).}
		\label{fig_conditional_span}
	\end{figure}

	Previous methods used to predict the start and end position of an answer span can be divided into two categories. The first one regards the generation of begin position and end position independently. We refer to this category as \textit{independent approach}. It can be written as $p_* = p\left(*|\mathbf{H}^*\right)$, where $*\in\{s, e\}$, the $s$ and $e$ denote start and end, respectively. $\mathbf{H}^*$ is the hidden representation, in which $\mathbf{H}^s$ and $\mathbf{H}^e$ usually have shared features. The other one constructs a dependent route from the start position when predicting the end position. We refer to this category as \textit{conditional approach}. It can be formalized as $p_s = p\left(s|\mathbf{H}^s\right),p_e = p\left(e|s,\mathbf{H}^e\right)$. This category usually reuses the predicted position information (e.g., $s$) to assist in the subsequent prediction. The difference between these two approaches is that the conditional approach considers the relationship between start and end positions, but the independent approach does not. In the literature, AMANDA \cite{Kundu2018a}, QANet \cite{Yu2018}, and SEBert \cite{Keskar2019} can be regarded as the independent approach, where the probabilities of the start and end positions are calculated separately with different representations. DCN \cite{Xiong2017}, R-NET \cite{Wang2017a}, BiDAF\footnote{We classify BiDAF as a conditional approach by its official implementation: \url{https://github.com/allenai/bi-att-flow}} \cite{Seo2017}, Match-LSTM \cite{Wang2017}, S-Net \cite{Tan2017}, SDNet \cite{Zhu2018}, and HAS-QA \cite{Pang2019} belong to the conditional approach. The probabilities are generated in sequence.

	The conditional approach empirically has an advantage over the independent approach. However, the output distributions of the previous conditional approaches are two probability vectors. It ignores some more possible start-end pairs. As an extension, every possible start (or end) position should have an end (or start) probability vector. Thus, the output conditional probabilities is a matrix.

	We propose a \textbf{Ma}trix-based \textbf{P}rediction approach (MaP) based on the above consideration in this paper. As Figure \ref{fig_conditional_span} shown, the key point is to consider as many probabilities as possible in \emph{training} and \emph{inference} phases. Specifically, we calculate a conditional probability matrix instead of a probability vector to expand the choices of start-end pairs. Because of more values contained in a matrix than a vector, there is a big challenge in the training phase of the MaP. That is the high computational cost and memory issues if the input sequence is long. As an instance, the matrix contains $262,144$ probability values if the sequence length is 512. Therefore, we propose a sampling-based training strategy to speed up the training and reduce the memory cost.

	The main contributions of our work are four-fold.
	\begin{itemize}
		\item A novel conditional approach is proposed to address the limitation of the probability vector generated by the vector-based conditional approach. It increases the likelihood of hitting the ground-truth start and end positions.
		\item A sampling-based training strategy is proposed to overcome the computation and memory issues in the training phase of the matrix-based conditional approach.
		\item An ensemble approach on both start-to-end and end-to-start directions of conditional probability is investigated to improve the accuracy of the answer span.
		\item We evaluate our strategy on SQuAD 1.1 and three other question answering benchmarks. The implementation of the matrix-based conditional approach is designed based on the BERT and BiDAF, which are the most competitive models, to test the generalization of our strategy. The consistent improvements in all datasets demonstrate the effectiveness of the strategy.
	\end{itemize}

	\section{Methodology}
	\label{sec_methodology}
	In this section, we first give the problem definition. Then we introduce a typical vector-based conditional approach. Next, we mainly introduce our matrix-based conditional approach and sampling-based training strategy. Finally, an ensemble approach on both start-to-end and end-to-start directions of conditional probability is discussed.
	
	\subsection{Problem Statement}
	\label{sec_problem}
	Given the passage $P=\{t_1, t_2, \cdots, t_n\}$ and the question $Q=\{q_1, q_2, \cdots, q_m\}$, the span extraction task needs to extract the continuous subsequence $A=\{t_{s}, \cdots, t_{e}\}$ ($1 \leq s \leq e \leq n$) from the passage as the right answer to the question, where $n$ and $m$ are the length of the passage and question respectively, $s$ and $e$ are the start and end position in the passage. Usually, the objective to predict $a=(s, e)$ is maximizing the conditional probability $p(a|P,Q)$.
	\begin{figure}[tbp]
		\centering
		\includegraphics[width=2.6in]{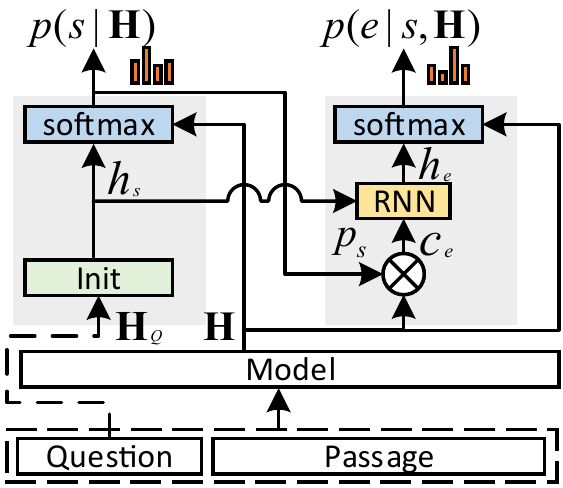}
		\caption{A typical implementation of the vector-based conditional approach.}
		\label{fig_pointer_net}
	\end{figure}

	\subsection{A Typical Vector-based Approach}
	We summarize a typical implementation of the vector-based conditional approach shown in Figure \ref{fig_pointer_net}. Previous mentioned R-NET, BiDAF, Match-LSTM, S-Net, and SDNet can be regarded as such implementation. Its backbone is the Pointer Network proposed by \citeauthor{Vinyals2015} (\citeyear{Vinyals2015}). The interactive representation $\mathbf{H} \in \mathbb{R}^{n \times d}$ between the given question $Q$ and passage $P$ is calculated as follows,
	\begin{align}
		\mathbf{H} = \mathfrak{M}(Q, P), \label{eq_model}
	\end{align}
	where $\mathfrak{M}$ is a neural network, e.g., Match-LSTM, QANet, BERT, and XLNet, $d$ is the dimension size of the representation. After generating the interactive representation, the next step is to predict the answer span.

	The main architecture of the span prediction is an RNN. As an instance, LSTM is used in \cite{Wang2017}, and GRU is adopted in \cite{Tan2017,Zhu2018}. Take the hidden representation $h_e \in \mathbb{R}^{k}$ of end position as an example, which is calculated as follows,
	\begin{align}
		h_e &= \textup{RNN}(h_s, c_e), \label{eq_rnn} \\
		c_e &= \mathbf{H}^{\top}p_s, \label{eq_ce}
	\end{align}
	where $p_s = p(s|\mathbf{H})$ is the start probability and $p_s \in \mathbb{R}^n$, $k$ is the dimension size of $h_e$. Then $p_e = p(e|s,\mathbf{H}) (p_e \in \mathbb{R}^n)$ can be calculated using $h_e$ as follows,
	\begin{align}
		p(e|s,\mathbf{H}) &= \textup{softmax} \Big(\mathbf{v}^\top \textup{tanh}\big(\mathbf{V}\mathbf{H}^{\top}+\llbracket{\mathbf{W}_eh_e}\rrbracket^n\big)\Big) \label{eq_h_e}
	\end{align}
	where $\llbracket{ \cdot }\rrbracket^n$ is an operation that generates a matrix by repeating the vector on the left $n$ times, $\mathbf{v} \in \mathbb{R}^{l}$, $\mathbf{V} \in \mathbb{R}^{l \times d}$, and $\mathbf{W}_e \in \mathbb{R}^{l \times k}$ are parameters to be learned.

	The calculation of $p(s|\mathbf{H})$ is similar to $p(e|s,\mathbf{H})$. The key is to obtain the hidden state $h_s$. A choice is to use an attention approach to condense the question representation into a vector. The process is as follows,
	\begin{align}
		p_{init} &= \textup{softmax} \Big(\mathbf{v}_Q^\top \textup{tanh}\big(\mathbf{V}_Q\mathbf{H}_Q^{\top}\big)\Big), \label{eq_att_pool} \\
		h_s &= \mathbf{H}_Q^{\top}p_{init}, \label{eq_h_s}
	\end{align}
	where $\mathbf{H}_Q \in \mathbb{R}^{m \times d}$ is the representation corresponding to $Q$, $\mathbf{v}_Q \in \mathbb{R}^{l}$, and $\mathbf{V}_Q \in \mathbb{R}^{l \times d}$ are parameters.

	There is a vast number of works on MRC. However, most of these works focus on the design of $\mathfrak{M}$ and generate the answer span based on the vector-based conditional approach. In this paper, we expand the vector to a probability matrix. Thus, many more possibilities can be covered. It is also a natural manner because that every start (or end) position should have an end (or start) probability vector.

	\subsection{Matrix-based Conditional approach}
\begin{figure}[tbp]
		\centering
		\includegraphics[width=3.0in]{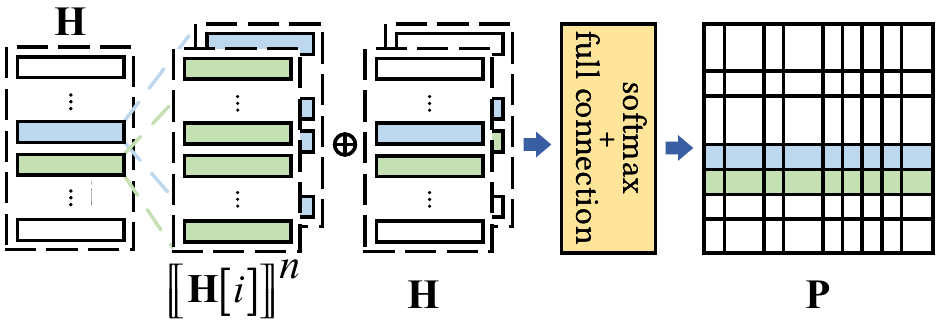}
		\caption{Matrix-based conditional approach.}
		\label{fig_matrix_process}
	\end{figure}

	As the previous description, the implementation of the vector-based conditional approach has a unified and important implementation step: create a `condition'. Take the forward direction (`condition' constructed from the start position to end position) of the vector-based conditional approach as an example, the `condition' is the probability vector $p_s$. The end probability vector $p_e$ can not be calculated until generating $p_s$. However, there is only one probability vector $p_e$ whatever the start position is. In this paper, we keep the `condition' step but propose calculating an individual $p_e$ for each start position. Specifically, the probability matrix $\mathbf{P}_e \in \mathbb{R}^{n \times n}$ is calculated as follows,
	\begin{align}
		\mathbf{P}_e^{(i)} &= \textup{softmax} \Bigg(\mathbf{v}^\top \textup{tanh}\bigg(\mathbf{V}\bigg[\mathbf{H}^{\top};\Big\llbracket{\big(\mathbf{H}[i]\big)^\top}\Big\rrbracket^n\bigg]\bigg)\Bigg) \label{eq_p}
	\end{align}
	where $\mathbf{P}_e^{(i)}$ denotes the $i$-th row of $\mathbf{P}_e$, $[ ; ]$ is a concatenate operation, $\llbracket{ \cdot }\rrbracket^n$ is an operation that generates a matrix by repeating the vector on the left $n$ times, $[i]$ means to choose the $i$-th row from the matrix $\mathbf{H}$, $\mathbf{v} \in \mathbb{R}^{l}$ and $\mathbf{V} \in \mathbb{R}^{l \times 2d}$ are parameters. Figure \ref{fig_matrix_process} illustrates the calculation process of Eq. (\ref{eq_p}).

	 Although the calculation is brief and can cover more probabilities than the vector-based approach, there is a big question on computation cost and memory occupation. The main computation cost comes from the matrix multiplication between $\mathbf{V}$ and $\bigg[\mathbf{H}^{\top};\Big\llbracket{\big(\mathbf{H}[i]\big)^\top}\Big\rrbracket^n\bigg]$ in Eq. (\ref{eq_p}), totally $n$ times such computation for $\mathbf{P}_e$. The number of probabilities is also $n$ times bigger than the vector-based conditional approach. It also causes the issue of out of memory (OOM), especially with a big $n$, due to intermediate gradient values needing cache in the training phase. We propose a sampling-based training strategy to solve the above issues.

	\subsection{Sampling-based Training Strategy}
	In order to train the probability matrix effectively, we propose a sampling-based strategy in the training phase. Given the hyper-parameter $k$, we first choose the indexes $\mathcal{\hat{I}}$ of top $k-1$ possibilities from $p_s^{(\textup{-}\hat{s})}$,
	\begin{align}
		\mathcal{\hat{I}} = \textup{top}\Big(p_s^{(\textup{-}\hat{s})}, k-1\Big), \label{eq_sample_start}
	\end{align}
	where $\textup{top}(p,\mathit{v})$ is an operation used to get the indexes of top $\mathit{v}$ values in $p$, $p^{(\textup{-}w)}$ contains all but $w$-th value of $p$, and $\hat{s}$ is the truth start position used as the supervised information in the training phase. Then, the $\hat{s}$ must merge to $\mathcal{\hat{I}}$,
	\begin{align}
		\mathcal{I} = \mathcal{\hat{I}} + \{\hat{s}\}, \label{eq_merge_start}
	\end{align}
	where $\mathcal{I}$ contains $k$ indexes.

	Eq. (\ref{eq_sample_start}) and Eq. (\ref{eq_merge_start}) promise that the sampled start probabilities must contain and only contain the target probability which we need to train in each iteration. The target probability is the $\hat{s}$-th value in $p_s$, and the bigger, the better.

\begin{figure}[tbp] 
		\centering
		\includegraphics[width=3.0in]{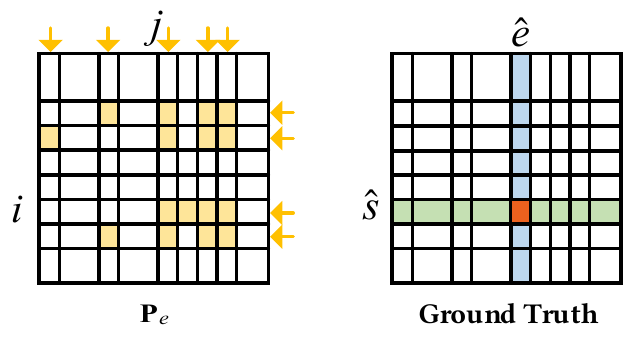}
		\caption{A sampling of probability matrix. Left: the calculated probability matrix with sampled top four positions (in both row and column directions colored with yellow background). Right: the ground truth matrix, where position $(\hat{s}, \hat{e})$ with the red background has probability 1.}
		\label{fig_choose_topK}
	\end{figure}

	After sampling the start probability vector, the computation cost of $\mathbf{P}_e$ decrease. For each $i \in \mathcal{I}$, executing Eq. (\ref{eq_p}) repeatedly can generate a sampling-based end probability matrix. It is noted that this sampling-based matrix is a part of the original $\mathbf{P}_e$. We refer to it as $\mathbf{\tilde{P}}_e$, and $\mathbf{\tilde{P}}_e \in \mathbb{R}^{k \times n}$. It is still a big issue of computation cost and memory occupation for $\mathbf{\tilde{P}}_e$ with a long sequence. So, we carry out similar operations in Eq. (\ref{eq_sample_start}) and Eq. (\ref{eq_merge_start}) for each row of $\mathbf{\tilde{P}}_e$ using $\hat{e}$ instead of $\hat{s}$, where $\hat{e}$ is the end truth position. Finally, the sampling-based matrix $\mathbf{\hat{P}}_e \in \mathbb{R}^{k \times k}$ is generated. It is small enough to train compared with $\mathbf{P}_e$. Figure \ref{fig_choose_topK} shows the sampling results colored with a yellow background on the left and corresponding ground truth matrix on the right.

	\subsection{Training}
	In the training phase, the objective function is to minimize the cross-entropy error averaged over start and end positions,
	\begin{align}
		\mathcal{L} &= \frac{1}{2}(\mathcal{L}_s + \mathcal{L}_e), \label{eq_all_loss} \\
		\mathcal{L}_s &= -\frac{1}{N}\sum_{i=1}^N \left( \mathbb{I} \left(\hat{s} \right) \big( \text{log} \left(p_s \right) \big)^\top \right), \\
		\mathcal{L}_e &= -\frac{1}{N}\sum_{i=1}^N \bigg( \mathcal{T} \big(\mathbb{I} \left(\hat{s}, \hat{e} \right)\big) \Big( \text{log} \left(\mathcal{T} ( \mathbf{\hat{P}}_e ) \right) \Big)^\top \bigg),  \label{eq_loss_e}
	\end{align}
	where $N$ is the number of data, $\mathbb{I}(\hat{s})$ means the one-hot vector of $\hat{s}$, $\mathbb{I} \left(\hat{s}, \hat{e} \right)$ means a zero matrix with a value of 1 in row $\hat{s}$ and column $\hat{e}$, and $\mathcal{T}()$ is a row wise flatten operation. The flatten operation makes the loss function on matrix-based distribution similar to that on vector-based distribution.

	As the introduction of the sampling-based training strategy, there are limited end probabilities that could be trained in each iteration. The extreme situation is $k$ equals to $n$, which makes all probability matrix calculate each time. As our previous argumentation, it is almost impossible for time and memory limitations. However, there is a question of what makes sampling strategy works. The following content gives some explanation based on gradient backpropagation.

	The gradient of the cross-entropy $\mathcal{L}_*$ to the predicted logits $z_*$ is,
	\begin{equation}
		\frac{\partial\mathcal{L}_*}{\partial z_*} =
			\begin{cases}
				p_*^{(i)} - 1, & \textup{if } i \textup{ is the ground-truth}; \\
				p_*^{(j)}, & others
				\end{cases}
		\label{eq_gradient}
	\end{equation}
	where $p_*=\textup{softmax}(z_*)$ is probabilities in which values are between 0 and 1 (exclusion). Thus $p_*^{(i)} - 1$ is negative, and $p_*^{(j)}$ is positive in most cases. As the parameters $\theta$ update usually follows $\theta_t=\theta_{t-1}-\eta \cdot \nabla_\theta\mathcal{L}(\theta)$ and learning rate $\eta$ is a positive value, the probability in ground-truth position should go up, and the probabilities in other sampled positions should go down.

\begin{figure}[tbp]
		\centering
		\includegraphics[width=3.0in]{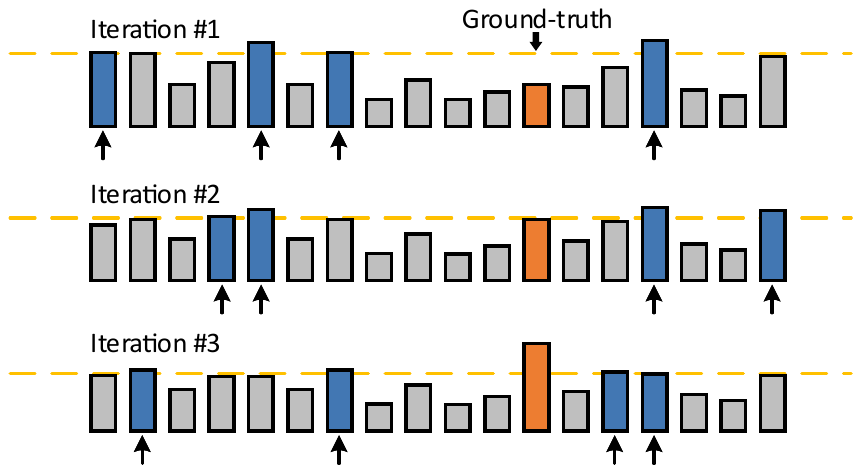}
		\caption{Sampling-based probabilities training ($k=5$). Block with red color is the ground-truth, blocks with blue color are the sampled probabilities. Probabilities with a gray background will not change their values in each iteration.}
		\label{fig_sampling_training_process}
	\end{figure}

	Figure \ref{fig_sampling_training_process} illustrates the sampling-based training process, where the parameter $k$ is set to 5. It means that there are extra top-4 probabilities (blue background) except ground-truth (red background) will be chosen to calculate. With the iteration going from \#1 to \#3, the probability in ground-truth position goes up, and that in sampled top-4 positions goes down. Such a sampling-based training approach has the same goal with the training on the whole probabilities, thus should have proximity results.

	\subsection{Ensemble for Inference}
	The vector-based conditional approach usually searches the span $(s,e)$ via the computation of $p_s^{(i)} \times p_e^{(j)}$ under the condition of $i \leq j$, and choices the $(i^*, j^*)$ with the highest $p_s^{(i^*)} \times p_e^{(j^*)}$ as the output in the inference phase. The matrix-based conditional approach follows the same idea, but the calculation of the probability is $p_s^{(i)} \times \mathbf{P}_e^{(i,j)}$ instead of $p_s^{(i)} \times p_e^{(j)}$. The $p_s^{(i)}$ is the $i$-th probability in $p_s$, and $\mathbf{P}_e^{(i,j)}$ is the probability in row $i$, column $j$ of $\mathbf{P}_e$.

	The above inference strategy only involves one direction, e.g., start-to-end direction (generate start position firstly, then generate end position), which is the most cases in previous works. An ensemble of both start-to-end and end-to-start directions is a good choice to improve the performance. The difference in end-to-start direction is that Eqs. (\ref{eq_p}-\ref{eq_loss_e}) should be repeated in the opposite direction. In other words, the start is replaced by $e$, and the end is replaced by $s$. Totally, there are two groups of probabilities, $(p_s, \mathbf{P}_e)$ and $(p_e, \mathbf{P}_s)$. In this paper, we design a type of ensemble strategy, which first chooses top $k$ pairs $\mathbf{F}=\{(i_f,j_f)\}$ with highest probability $p_s^{(i_f)} \times \mathbf{P}_e^{(i_f,j_f)}$, then chooses top $k$ pairs $\mathbf{B}=\{(j_b,i_b)\}$ with highest probability $p_e^{(j_b)} \times \mathbf{P}_s^{(j_b,i_b)}$. It is noted that some pairs may have the same position, e.g., $(3_f, 5_f)$ and $(5_b, 3_b)$. If there are the same elements, we prune away them in $\mathbf{B}$. Then, we choose the $(i^*, j^*)$ with highest probability in $\mathbf{F} \cup \mathbf{B}$.

	The overall training procedure of MaP is summarized in Algorithm \ref{alg_mpv}.
	\begin{algorithm}[tp]
		\caption{MaP Training Algorithm}
		\label{alg_mpv}
		\begin{algorithmic}[1]
		\REQUIRE $N$ pairs of passage $P$ and question $Q$, $k$ used to choose top probabilities;
		\ENSURE Learned MaP model
		\STATE Initialize all learnable parameters $\Theta$;
		\REPEAT
		\STATE Select a batch of pairs from corpus;
		\FOR{each pair $(P, Q)$}
		\STATE Use a neural network $\mathfrak{M}$ to generate the representation $\mathbf{H}$; \COMMENT{(Eq.\;\ref{eq_model})}
		\STATE Compute start probability vector $p_s$; \COMMENT{(Eqs.\;\ref{eq_h_e}-\ref{eq_h_s})}
		\STATE Sample indexes $\mathcal{I}$ by choosing top $k-1$ probabilities of $p_s$; \COMMENT{(Eqs.\;\ref{eq_sample_start},\ref{eq_merge_start})}
		\STATE Compute end probability matrix $\mathbf{P}_e$; \COMMENT{(Eq.\;\ref{eq_p})}
		\STATE Compute objective $\mathcal{L}$; \COMMENT{(Eq.\;\ref{eq_all_loss}-\ref{eq_loss_e})}
		\ENDFOR
		\STATE Use the backpropagation algorithm to update parameters $\Theta$ by minimizing the objective with the batch update mode
		\UNTIL{\emph{stopping criteria is met}}
		\end{algorithmic}
	\end{algorithm}

	\section{Experiments}
	In this section, we conduct experiments to evaluate the effectiveness of the proposed MaP.
	\begin{table*}[thbp]
		\begin{center}
			\setlength{\tabcolsep}{2.6mm}{
				\begin{tabular}{|ll|cc|cc|cc|cc|}
					\hline
					\multicolumn{2}{|c|}{\multirow{2}{*}{Models}} & \multicolumn{2}{c|}{SQuAD}      & \multicolumn{2}{c|}{NewsQA}     & \multicolumn{2}{c|}{HotpotQA}   & \multicolumn{2}{c|}{Natural Questions} \\
					\multicolumn{2}{|c|}{} & EM             & F1             & EM             & F1             & EM             & F1             & EM                 & F1                \\ \hline \hline
					BERT-Base              & InD                & 81.24          & 88.38          & 52.59          & 67.12          & 59.01          & 75.69          & 67.31              & 78.96             \\
										   & \textbf{MaP}$_{F}$     & 81.78          & 88.59          & 52.66          & 66.50          & 59.82          & 75.81          & 67.68              & 78.99             \\
										   & \textbf{MaP}$_{E}$ & \textbf{82.12} & \textbf{88.63}          & \textbf{53.06} & \textbf{67.37} & \textbf{60.55} & \textbf{76.12} & \textbf{68.21} & \textbf{79.09}             \\ \hline \hline
					BERT-Large             & InD                & 84.05          & 90.85          & 54.46          & 69.61          & 62.26          & 78.18          & 69.44              & 80.93             \\
										   & \textbf{MaP}$_{F}$     & 84.50          & 90.89          & 54.84          & 68.73          & 63.19          & 78.99          & 69.56              & 80.49             \\
										   & \textbf{MaP}$_{E}$         & \textbf{84.79} & \textbf{90.89} & \textbf{55.29} & \textbf{69.98} & \textbf{63.70} & \textbf{79.25} & \textbf{69.91}              & \textbf{81.22}             \\ \hline \hline
					BiDAF                  & VCP              & 68.57          &  78.23 & 44.04          & 58.07 & 47.31          & 62.42          & 56.95              & 68.79             \\
										   & \textbf{MaP}$_{F}$     & 68.85          & 78.06          & 44.19          & 58.65          & 50.25          & 65.21          & 57.04              & 68.87             \\
										   & \textbf{MaP}$_{E}$ & \textbf{69.55} & \textbf{78.91}          & \textbf{44.25} & \textbf{58.91}          & \textbf{51.45} & \textbf{66.74} & \textbf{57.21}     & \textbf{69.08}    \\ \hline
				\end{tabular}
			}
		\end{center}
		\caption{\label{table_results} The performance (\%) of EM and F1 on SQuAD 1.1 and three MRQA extractive question answering tasks. MaP$_{F}$ is the matrix-based conditional approach calculating on start-to-end direction. MaP$_{E}$ means the ensemble of both directions of matrix-based conditional approach. InD denotes the independent approach. VCP is vector-based conditional approach.}
	\end{table*}

	\subsection{Datasets}
	We first evaluate our strategy on SQuAD 1.1, which is a reading comprehension benchmark. The benchmark benefits to our evaluation compared with its augmented version SQuAD 2.0 due to its questions always have a corresponding answer in the given passages. We also evaluate our strategy on three other datasets from the MRQA 2019 Shared Task\footnote{\url{https://github.com/mrqa/MRQA-Shared-Task-2019}}: NewsQA \cite{Trischler2017}, HotpotQA \cite{Yang2018}, Natural Questions \cite{Kwiatkowski2019}. As the SQuAD 1.1 dataset, the format of the task is extractive question answering. It contains no unanswerable or non-span answer questions. Besides, the fact that these datasets vary in both domain and collection pattern benefits for the evaluation of our strategy on generalization across different data distributions. Table \ref{table_dataset} shows the statistics of these datasets.
	\begin{table}[thbp]
		\begin{center}
			\begin{tabular}{|l|c|c|}
				\hline
				Dataset     & Training & Development \\ \hline 
				SQuAD 1.1            & 86,588 & 10,507 \\ 
				NewsQA           & 74,160 & 4,212 \\ 
				HotpotQA               & 72,928 & 5,904 \\ 
				Natural Questions               & 104,071 & 12,836 \\ \hline
				\end{tabular}
		\end{center}
		\caption{\label{table_dataset} The statistics of datasets.}
	\end{table}

	\subsection{Baselines}
	To validate the effectiveness and generalization of our proposed strategy on the span extraction, we implement it using two strong backbones, BERT and BiDAF. Specifically, we borrow their main bodies except the top layer to implement the proposed strategy to finish the span extraction on different datasets. Some more tests on other models, e.g., XLNet \cite{Yang2019a} and SpanBERT \cite{Joshi2019}, and datasets will be our future work.
	\begin{itemize}
		\item \textbf{BERT}: BERT is an empirically powerful language model, which obtained state-of-the-art results on eleven natural language processing tasks in the past \cite{Devlin2019}. The original implementation in their paper on the span prediction task belongs to the independent approach. Both BERT-base and BERT-large with uncased pre-trained weights are used in comparison to investigating the effect of the ability of language model on span extraction with different prediction approaches.
		\item \textbf{BiDAF}: BiDAF is used as a baseline of the vector-based conditional approach \cite{Seo2017}. The use of a multi-stage hierarchical process and a bidirectional attention flow mechanism makes its representation powerful.
	\end{itemize} 

	There are four strategies of span extraction involved in our comparison: \textbf{InD} denotes the independent approach; \textbf{VCP} is the vector-based conditional approach; \textbf{MaP}$_{F}$ is our matrix-based conditional approach calculating on start-to-end direction; \textbf{MaP}$_{E}$ means the ensemble of both directions of matrix-based conditional approach. The InD is used to compare with MaP$_{F}$ and MaP$_{E}$ in BERT, and the VCP is used to compare with MaP$_{F}$ and MaP$_{E}$ in BiDAF.

	\subsection{Experimental Settings}
	We implement the BERT and BiDAF following the official settings for a fair comparison. For the BERT, we train for 3 epochs with a learning rate of $5e\text{-}5$ and a batch size of 32. The max sequence length is 384 for SQuAD 1.1 and 512 for other datasets, and a sliding window of size 128 is used for all datasets is the sentence is longer than the max length. For the BiDAF, we keep all original settings except a difference that we use ADAM \cite{Kingma2015} optimizer with a learning rate of $1e\text{-}3$ in the training phase instead of AdaDelta \cite{Zeiler2012} for a stable performance. Following the work from \cite{Rajpurkar2016}, we evaluate the results using Exact Match (EM) and Macro-averaged F1 score. The sampling parameter $k$ is set to 20 for our strategy. We implement our model in python using the pytorch-transformers library\footnote{\url{https://github.com/huggingface/pytorch-transformers}} for BERT and the AllenNLP library\footnote{\url{https://github.com/allenai/allennlp}} for BiDAF. The reported results are average scores of 5 runs with different random seeds. All computations are done	on 4 NVIDIA Tesla V100 GPUs.

	\subsection{Main Results}
	The results of our strategies as well as the baselines are shown in Table \ref{table_results}. All these values come from the evaluation of the development sets in each dataset due to the test sets are withheld. Nevertheless, our strategy achieves a consistent improvement compared with the independent approach and the vector-based conditional approach. The values with a bold type mean the winner across all strategies. As we can observe, the MaP$_{E}$ wins 16 out of 16 in both BERT-base and BERT-large groups. It proves that the ensemble of both directions is helpful for the span extraction. In the BiDAF group, The MaP$_{E}$ is also the best on all datasets compared with VCP. It shows the robustness of our matrix-based conditional approach in language models. The fact that the MaP$_{F}$ wins 12 out of 12 in EM, and 8 out of 12 in F1 demonstrates that the matrix-based conditional approach is capable of predicting a clean answer span that matches human judgment exactly. We suppose the reason is that more start-end position pairs considered in the probability matrix can enhance the interaction and constraint between the start and end, thus, make the MaP$_{F}$ perform more consistently in EM than in F1.
	\begin{figure}[thbp]
		\centering
		\includegraphics[width=3.0in]{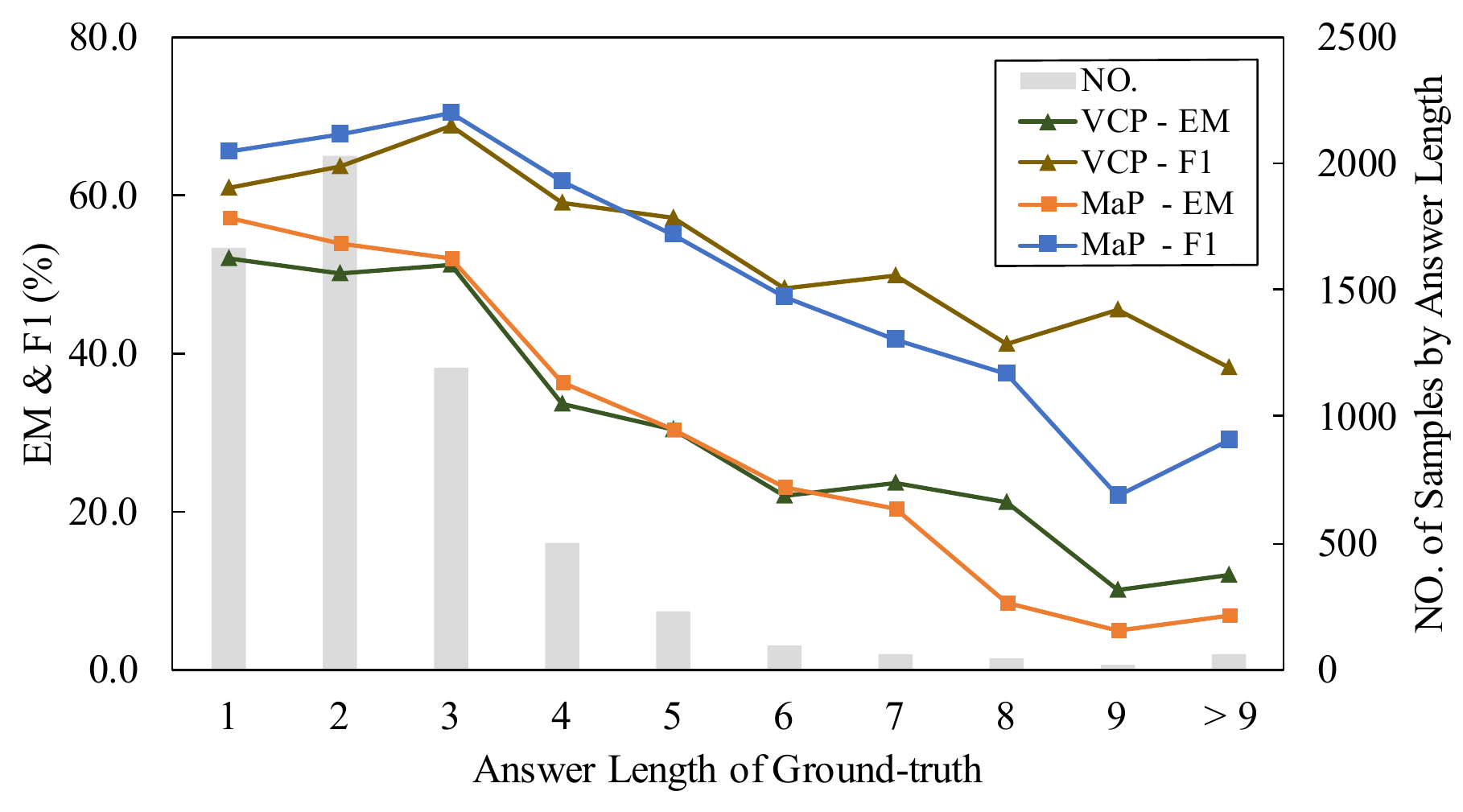}
		\caption{EM and F1 of MaP$_{F}$ and VCP based on BiDAF under different answer length.}
		\label{fig_results_on_answerlength}
	\end{figure}

	\subsection{Strategy Analysis}
	Figure \ref{fig_results_on_answerlength} shows how the performance changes with respect to the answer length, which is designed on HotpotQA. We can see that the matrix-based conditional approach works better than the vector-based conditional approach as the span decrease in length. Since the short answers have a high rate in all answer spans, so the matrix-based conditional approach is better for the answer span task. In other words, this observation supports the ensemble of both directions as $_{E}$ does. The MaP$_{E}$ combining the MaP$_{F}$'s advantage in short answers and the VCP's advantage in long answers can get a better result than any of them.
	\begin{figure}[thbp]
		\centering
		\includegraphics[width=3.0in]{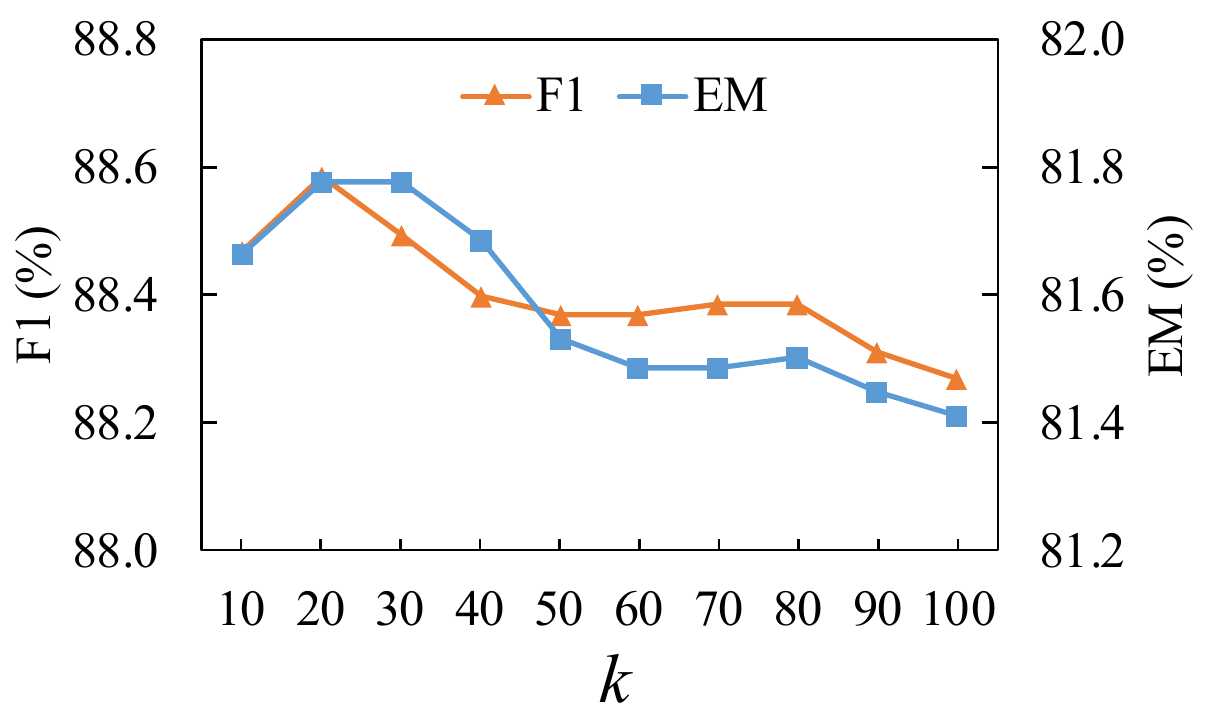}
		\caption{Impact of hyper-parameter $k$ in MaP$_{F}$ on SQuAD 1.1 with BERT-base as the backbone.}
		\label{fig_impact_of_k}
	\end{figure}

	We investigate the impact of $k$ used to choose the top probabilities in the training phase. The results are shown in Figure \ref{fig_impact_of_k}. With the increase of $k$, the EM and F1 show a downtrend. The best performance happens at $k=20$. We guess that choosing more probabilities makes the training difficult and brings extra noises to the candidate positions. E.g., if $k$ is set to 30, the number of candidate probabilities will be 900, which is larger than the sequence length 512 in vector-based conditional approach.
	\begin{figure}[thbp]
		\centering
		\includegraphics[width=3.0in]{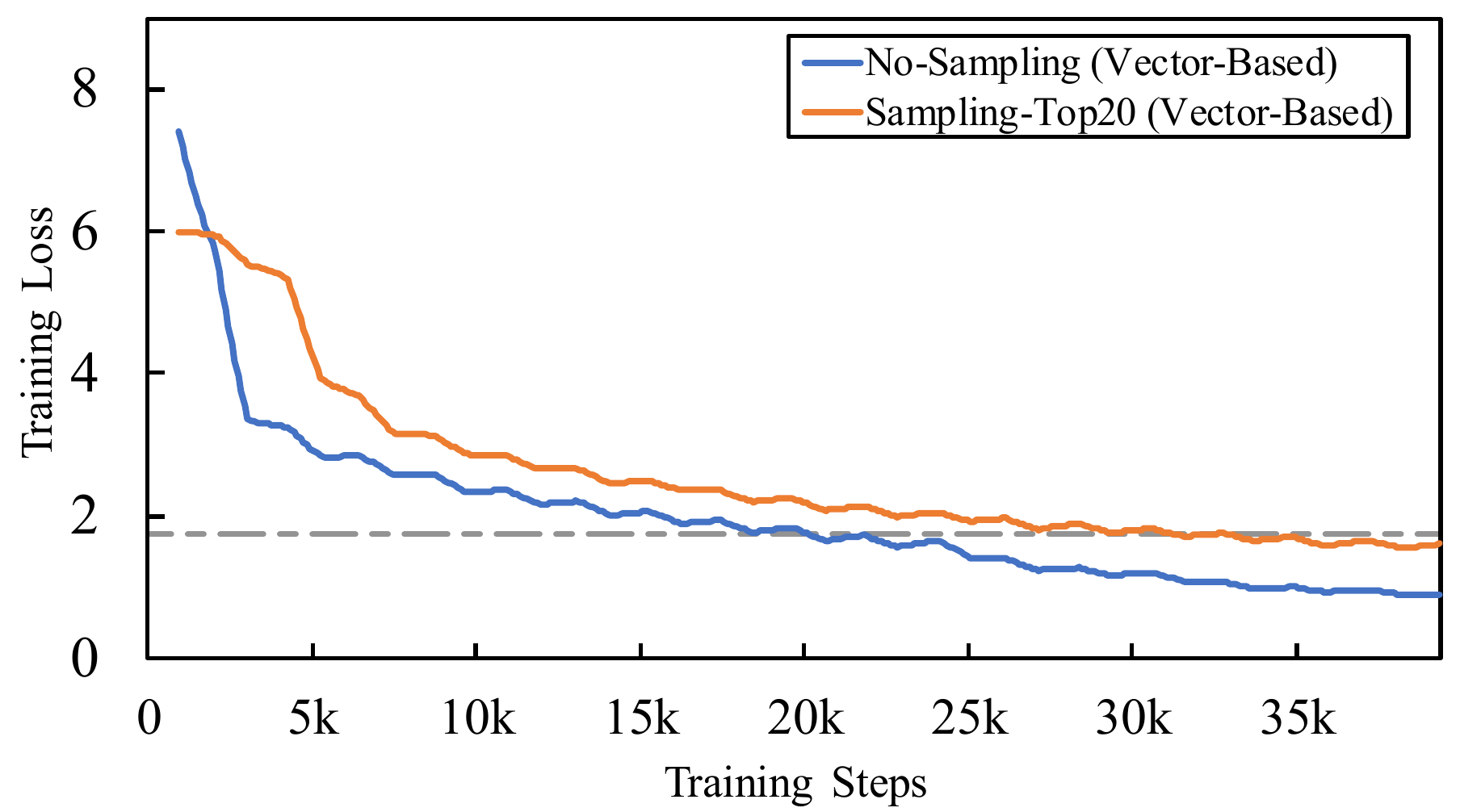} 
		\caption{Convergence of sampling-based training strategy on BERT.}
		\label{fig_convergence_training_process}
	\end{figure}

	We analyze the convergence of the sampling-based training strategy on SQuAD 1.1. Due to the effectiveness of the sampling-based training strategy is proved in MaP, we conduct an further experiment under the VCP to prove its generalization. Figure \ref{fig_convergence_training_process} demonstrates the results. As our expectation, the sampling-based training strategy optimizes the model as training in whole samples. However, it will cost longer training steps to get the same loss compared with standard training. So our sampling-based training strategy is good for the training of the matrix-based conditional approach.

	\section{Related Work}
	\label{sec_related_work}
	Machine reading comprehension is an important topic in the NLP community. More and more neural network models are proposed to tackle this problem, including DCN \cite{Xiong2017}, R-NET \cite{Wang2017a}, BiDAF \cite{Seo2017}, Match-LSTM \cite{Wang2017}, S-Net \cite{Tan2017}, SDNet \cite{Zhu2018}, QANet \cite{Yu2018}, HAS-QA \cite{Pang2019}. Among various MRC tasks, span extraction is a typical task that extracting a span of text from the corresponding passage as the answer of a given question. It can well overcome the weakness that words or entities are not sufficient to answer questions \cite{Liu2019}.

	Previous models proposed for span extraction mostly focus on the design of architecture, especially on the representation of question and passage, and the interaction between them. There are few works devoted to the top-level design of span output, which refers to the probabilities generation from the representation. We divide the previous top-level design into two categories, independent approach and conditional approach. The independent approach is to predict the start and end positions in the given passage independently \cite{Kundu2018,Yu2018}. Although the independent approach has a simple assumption, it works well when the input features are strong enough, e.g., combining with BERT \cite{Devlin2019}, XLNet \cite{Yang2019}, and SpanBERT \cite{Joshi2019}. Nevertheless, since there is a kind of dependency relationship between start and end positions, the conditional approach has advancements over the independent approach.

	A typical work on the conditional approach comes from \citeauthor{Wang2017} (\citeyear{Wang2017}). They proposed two different models based on the Pointer Network. One is the sequence model which produces a sequence of answer tokens as the final output, and another is the boundary model which produces only the start token and the end token of the answer. The experimental results demonstrate that the boundary model (span extraction) is superior to the sequence model on both EM and F1. The R-NET \cite{Wang2017a}, BiDAF \cite{Seo2017}, S-Net \cite{Tan2017}, SDNet \cite{Zhu2018} have the same output layer and inference phase with the boundary model in \cite{Wang2017}. \citeauthor{Lee2016} (\citeyear{Lee2016}) presented an architecture that builds fixed length representations of all spans in the passage with a recurrent network to address the answer extraction task. The computation cost is decided by the max-length of the possible span and the sequence length. The experimental results show an improvement on EM compared with the endpoints prediction that independently predicts the two endpoints of the answer span.

	However, previous works related to the conditional approach are always based on a probability vector. We investigate another possible matrix-based conditional approach in this paper. Besides, a well-matched training strategy is proposed to our approach, and forward and backward conditional possibilities are also integrated to improve the performance.

	\section{Conclusion}
	In this paper, we first investigate different approaches of span extraction in MRC. To improve the current vector-based conditional approach, we propose a matrix-based conditional approach. More careful consideration of the dependencies between the start and end positions of the answer span can predict their values better. We also propose a sampling-based training strategy to address the training process of the matrix-based conditional approach. The final experimental results on a wide of datasets demonstrate the effectiveness of our approach and training strategy.
	
	\section*{Acknowledgments}
	This work was supported by National Key R\&D Program of China (2019YFB2101802) and Sichuan Key R\&D project (2020YFG0035).

	\bibliographystyle{acl_natbib}
	\bibliography{Ref-QASpanExtraction}
\end{document}